\documentclass{article}
\usepackage{spconf,amsmath,graphicx,hyperref,multirow}

\newcommand{\std}[1]{{\scriptsize$\pm$#1}}

\title{EviDETR: Preserving Query-Relevant Temporal Evidence for Moment Retrieval and Highlight Detection}

\name{
Haoran Sun$^{*}$,
Yufan Li$^{*}$,
Qichen Zhang$^{*}$,
Haoran Zhao,
Shuqi Wang
\thanks{*These authors contributed equally to this work.
\quad Project page: \url{https://kevin-kas.github.io/Evi-detr-webpage/}}
}

\address{
Beijing Normal-Hong Kong Baptist University, Zhuhai, China\\
\{t330034045,t330034029,t330034072,t330026229\}@mail.bnbu.edu.cn\\
shuqiwhat@gmail.com
}
\begin{document}
%
\maketitle
\begin{abstract}
Joint video moment retrieval and highlight detection requires identifying query-relevant temporal segments while estimating clip-level saliency, yet DETR-style pipelines do not explicitly preserve query-relevant evidence throughout encoding, decoding, and cross-task prediction. We propose \textbf{EviDETR}, an evidence-preserving framework with three components. Semantic-aware Feature Reweighting (SFR) enhances query-relevant clip representations through saliency estimation and cross-modal interaction. A Temporal Top-2 Mixture-of-Experts (TTop2MoE) decoder performs query-adaptive refinement via sparse expert routing. MR-to-HD (MR2HD) fusion transfers span-level retrieval evidence to clip-level highlight prediction through confidence-weighted multi-scale aggregation. Using CLIP+SlowFast features, EviDETR achieves 69.29 R1@0.5, 54.77 R1@0.7, and 48.41 Avg. mAP for moment retrieval on QVHighlights, together with 41.83 HD-mAP and 68.33 HIT@1. Strong results on TACoS and Charades-STA further demonstrate cross-dataset transferability.

\end{abstract}
\section{INTRODUCTION}
\label{sec:intro}

\begin{figure*}[t]
    \centering
    \begin{minipage}[c]{0.44\textwidth}
        \centering
        \textbf{(a)}\par\vspace{1pt}
        \includegraphics[width=\linewidth]{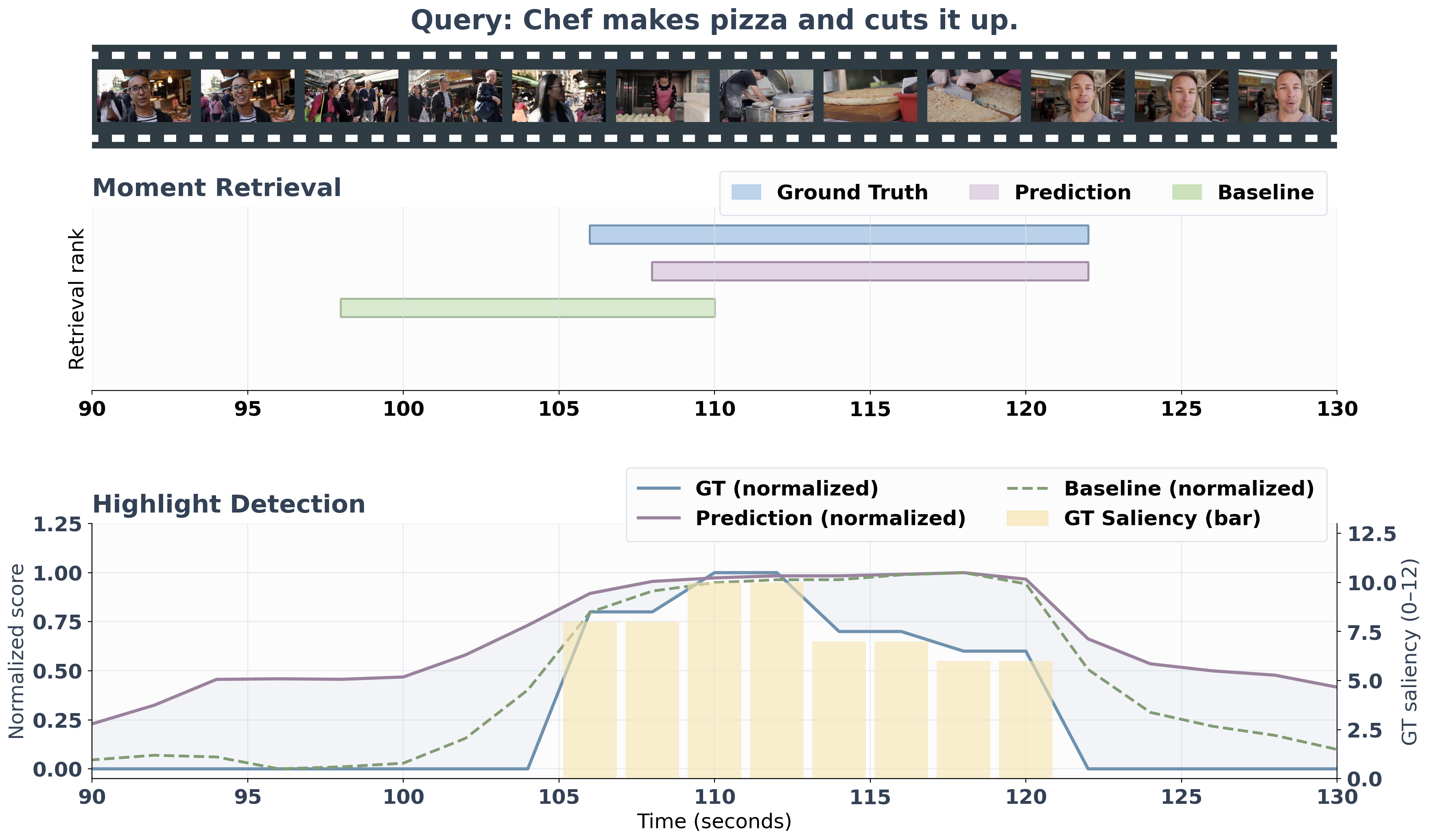}
    \end{minipage}
    \hfill
    \begin{minipage}[c]{0.53\textwidth}
        \centering
        \textbf{(b)}\par\vspace{1pt}
        \includegraphics[width=\linewidth]{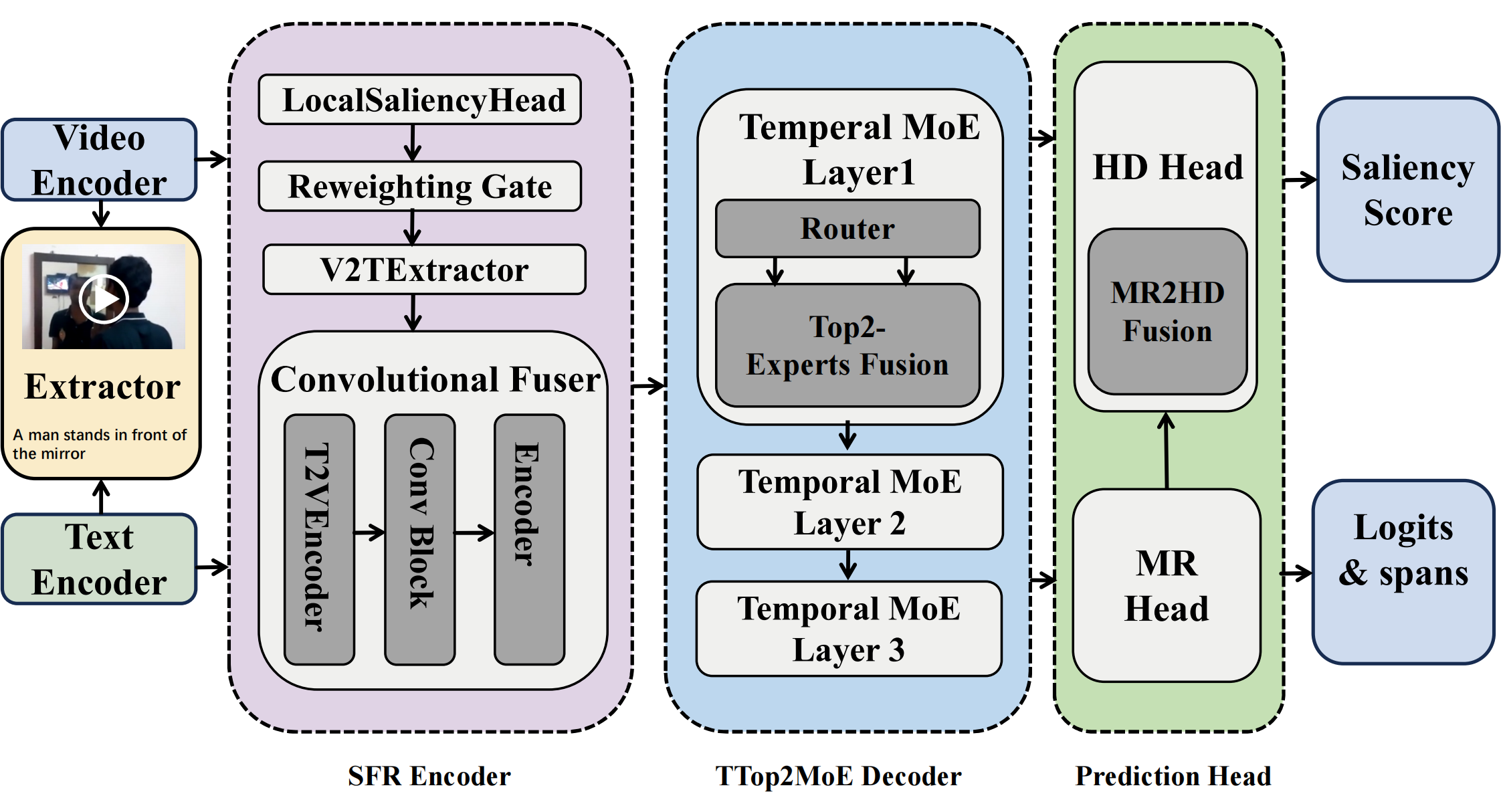}
    \end{minipage}
    \caption{Joint MR/HD task illustration and overview of EviDETR. (a) Given a natural-language query, MR localizes the query-relevant temporal segment, while HD estimates clip-level saliency. (b) EviDETR preserves query-relevant evidence through SFR encoding, query-adaptive TTop2MoE decoding, and MR2HD evidence fusion.}
    \label{fig:overview}
\end{figure*}
Video moment retrieval (MR) aims to localize the temporal segment described by a natural-language query, while highlight detection (HD) estimates the query-dependent saliency of individual video clips, as illustrated in Fig.~\ref{fig:overview}(a). Since both tasks require identifying temporally localized content that is semantically relevant to the query, recent studies have increasingly addressed MR and HD within a unified framework. In particular, DETR-style approaches formulate temporal grounding as an end-to-end set prediction problem and have achieved strong performance through query-dependent representation learning, temporal interaction, and cross-task modeling \cite{momentdetr,qddetr,trdetr,cgdetr}.

Recent methods tackle different bottlenecks in joint MR/HD.
CG-DETR~\cite{cgdetr} calibrates query dependency through
correlation-guided modeling, while BAM-DETR~\cite{bamdetr} emphasizes
boundary-aligned localization and TDP-DETR~\cite{tdpdetr} models
temporal dynamics. EviDETR instead focuses on how query-relevant
evidence is carried through the encoder--decoder--cross-task pipeline,
from relevance enhancement before decoding to conditional refinement
and explicit MR-to-HD evidence transfer.

Despite these advances, existing DETR-style joint MR/HD pipelines lack explicit mechanisms for preserving query-relevant temporal evidence across several stages of the pipeline. First, during encoding, relevant clips are mixed with large amounts of irrelevant video content, motivating explicit query-conditioned emphasis before decoding. Second, standard Transformer decoders apply the same feed-forward transformation to every temporal query, even though different moment hypotheses benefit from different refinement functions. A single shared FFN therefore applies a common transformation across temporal queries without explicit query-specific conditional computation. Third, although MR and HD are optimized jointly, span-level localization evidence from the MR branch is often only indirectly connected to clip-level saliency prediction, leaving useful retrieval cues under-exploited \cite{trdetr,uvcom}.

To address these limitations, we propose \textbf{EviDETR}, an evidence-preserving DETR framework for joint moment retrieval and highlight detection, as illustrated in Fig.~\ref{fig:overview}. First, \textbf{Semantic-aware Feature Reweighting (SFR)} estimates local query--clip relevance and enhances query-relevant video representations through semantic reweighting and cross-modal interaction before decoding. Second, the \textbf{Temporal Top-2 Mixture-of-Experts (TTop2MoE)} decoder dynamically routes each temporal query to a sparse subset of experts. This conditional computation allows temporal queries to activate selected parameter subsets, while Top-2 aggregation combines two routed expert transformations for conditional refinement. Third, the \textbf{MR-to-HD (MR2HD)} module explicitly transfers span-level localization evidence from MR to HD through multi-scale temporal aggregation, allowing retrieved moment cues to directly guide clip-level saliency estimation.

We evaluate EviDETR on QVHighlights, TACoS, and Charades-STA using CLIP+SlowFast features. On QVHighlights, EviDETR achieves 69.29 R1@0.5, 54.77 R1@0.7, and 48.41 Avg.\ mAP for MR, together with 41.83 HD-mAP and 68.33 HIT@1 for HD. The model also shows strong performance on TACoS and Charades-STA. Our main contributions are:
\begin{itemize}
    \item We formulate three evidence-preservation targets in DETR-style joint MR/HD: query-relevant emphasis during encoding, query-specific adaptation during refinement, and explicit MR-to-HD evidence transfer.
    \item We propose SFR, TTop2MoE, and MR2HD to respectively enhance query-relevant representations, introduce sparse Top-2 expert routing for conditional refinement, and explicitly transfer localization evidence from MR to HD.
    \item Under the CLIP+SlowFast feature setting, EviDETR achieves leading or second-best results across all reported QVHighlights metrics and strong cross-dataset performance on TACoS and Charades-STA.
\end{itemize}

\section{Proposed Method}

Given clip-level video features $V=\{v_t\}_{t=1}^{T}$ and
token-level query features $Q=\{q_n\}_{n=1}^{N}$, we project both
modalities into a shared $d$-dimensional space. As illustrated in
Fig.~\ref{fig:overview}, EviDETR consists of three components: Semantic-aware Feature
Reweighting (SFR), a Temporal Top-2 Mixture-of-Experts (TTop2MoE)
decoder, and MR-to-HD (MR2HD) evidence fusion. SFR enhances
query-relevant temporal representations before decoding, TTop2MoE
enables query-adaptive refinement through sparse expert routing, and
MR2HD transfers moment-level localization evidence to clip-level
highlight prediction.

\subsection{Semantic-aware Feature Reweighting}

In long videos, query-relevant clips can be obscured by large amounts
of irrelevant temporal content. SFR therefore estimates query--clip
relevance before deeper temporal modeling and uses it to emphasize the
most relevant clips.
Given a pooled query representation $g_q=\mathrm{Pool}(Q)$, we
compute
\begin{equation}
a_t=\sigma\!\left(\psi_{\mathrm{sal}}([v_t;g_q])\right),
\qquad
v_t^{a}=a_t v_t,
\end{equation}
where $a_t$ measures the local relevance of clip $t$ to the query.
To retain fine-grained cross-modal information, the reweighted video
features further interact with query tokens through a tri-linear
similarity:
\begin{equation}
e_{t,n}=w_v^\top v_t^{a}+w_q^\top q_n
+w_m^\top(v_t^{a}\odot q_n).
\end{equation}
Bidirectional video--text attention is then computed from
$e_{t,n}$ and fused with the reweighted video features. The resulting
query-aware sequence is temporally encoded to form the memory
$M=\{m_t\}_{t=1}^{T}$ for subsequent decoding.

\subsection{Temporal Top-2 Mixture-of-Experts Decoder}

Standard Transformer decoders apply the same FFN to every temporal query. However, different moment hypotheses exhibit different refinement needs, while a shared FFN does not explicitly adapt its transformation to individual temporal queries. Inspired by sparse conditional computation\cite{shazeer2017moe,gshard}, we replace the shared FFN with a Top-2 MoE module.

For the $i$-th query representation $\tilde z_i^l$ after
self-attention and cross-attention at decoder stage $l$, the router
predicts
\begin{equation}
\pi_i^l=\mathrm{Softmax}(W_g^l\tilde z_i^l),
\qquad
\mathcal{E}_i^l=\mathrm{TopK}(\pi_i^l,2).
\end{equation}
The selected experts are combined according to their normalized
routing probabilities:
\begin{equation}
\mathrm{MoE}(\tilde z_i^l)
=
\sum_{e\in\mathcal{E}_i^l}
\alpha_{i,e}^l\,\mathrm{Expert}_e^l(\tilde z_i^l),
\end{equation}
where
$\alpha_{i,e}^{l}=\pi_{i,e}^{l}/\sum_{e'\in\mathcal{E}_i^l}\pi_{i,e'}^{l}$.
The router therefore allows temporal queries to activate selected
parameter subsets. Unlike hard Top-1 routing, Top-2 routing
combines two expert transformations, enabling complementary refinement
while retaining sparse conditional computation. We employ three
sequential decoder stages for progressive temporal refinement.

\subsection{MR-to-HD Evidence Fusion}

The MR head predicts relevant moments
$\hat{\mathcal{M}}=\{(\hat s_i,\hat e_i,\hat p_i)\}_{i=1}^{N_q}$,
where $\hat p_i$ denotes foreground confidence. MR2HD explicitly
uses these spans to guide HD. We first encode the highest-confidence
span with a GRU and use its dot-product response with the SFR-enhanced
clip features $v_t^{a}$ to reweight the clip memory, yielding
$m_t^{(0)}$. We then extract span evidence at temporal scales
$\mathcal{A}=\{1,2,4\}$. A 1D adaptation of ROIAlign \cite{maskrcnn}
converts each predicted span into a fixed-length representation
$z_i^a$, and candidate spans are confidence-weighted:
\begin{equation}
z_{\mathrm{MR}}
=
\sum_{a\in\mathcal{A}}\lambda_a
\sum_{i=1}^{N_q}
\frac{\exp(\kappa\hat p_i)}
{\sum_j\exp(\kappa\hat p_j)}
z_i^a.
\end{equation}
The scale weights $\lambda_a$ are learned via softmax and initialized
uniformly. We parameterize $\kappa=\exp(\tau)$ with learnable
$\tau_0=3$. The aggregated evidence is transferred to clips by
\begin{equation}
\rho_t=\mathrm{Cos}(v_t^{a},z_{\mathrm{MR}}),
\qquad
\tilde m_t=m_t^{(0)}
+\lambda_{\mathrm{fuse}}\rho_t m_t^{(0)},
\end{equation}
where $\lambda_{\mathrm{fuse}}=\tanh(s)$ is learnable with $s_0=0$.
The fused memory $\tilde m_t$ is used for highlight prediction.

\subsection{Training Objective}

Following DETR-style set prediction \cite{detr}, MR predictions are
matched to ground truth using Hungarian matching and optimized with
foreground classification, $\ell_1$ span regression, and generalized
IoU (GIoU) loss \cite{giou}. We apply the same MR loss to the two
intermediate decoder stages:
\begin{equation}
\mathcal{L}
=
\mathcal{L}_{\mathrm{task}}
+
\sum_{l=1}^{L-1}\mathcal{L}_{\mathrm{MR}}^{l}.
\end{equation}
The intermediate MR loss is defined as
\begin{equation}
\mathcal{L}_{\mathrm{MR}}^{l}
=
4\mathcal{L}_{\mathrm{cls}}^{l}
+
10\mathcal{L}_{1}^{l}
+
\mathcal{L}_{\mathrm{giou}}^{l}.
\end{equation}
Here, $\mathcal{L}_{\mathrm{task}}$ denotes the final joint MR/HD
objective. For QVHighlights, we use
$(\lambda_{\mathrm{cls}},\lambda_{1},\lambda_{\mathrm{giou}})
=(4,10,1)$.
The saliency, multimodal-alignment, and MoE load-balancing weights
are $1.0$, $0.6$, and $0.02$, respectively. We use $\mathcal{L}_{\mathrm{bal}} = E \sum_{e=1}^{E} \bar{p}_e f_e$,
where $\bar{p}_e$ and $f_e$ denote the mean routing probability and
Top-2 assignment frequency, respectively, averaged over the three decoder stages.
Deep supervision provides direct optimization signals to earlier decoder stages.

\section{Experiments}
\subsection{Experimental Setup}

\textbf{Datasets and metrics.}
We evaluate EviDETR on three widely used temporal grounding
benchmarks: QVHighlights \cite{momentdetr},
Charades-STA \cite{tall}, and TACoS \cite{tacos}.
QVHighlights jointly evaluates moment retrieval (MR) and highlight
detection (HD), while Charades-STA and TACoS mainly evaluate
temporal moment localization. For QVHighlights, we report
R1@0.5, R1@0.7, and average mAP for MR, together with HD-mAP
and HIT@1 for HD. For Charades-STA and TACoS, we report
R1@0.5 and R1@0.7. QVHighlights main and ablation results use the
validation split. In Table~\ref{tab:cross_dataset}, EviDETR uses the
Charades-STA test split and TACoS test split.

\textbf{Implementation details.}
For all three datasets, we use CLIP+SlowFast features, with CLIP
\cite{clip} and SlowFast \cite{slowfast} for video representation and
CLIP for text queries. The hidden dimension is 256 and the number of
temporal queries is 10. TTop2MoE has three decoder stages, eight
experts per stage, and Top-2 routing; each expert uses FFN width 1024.
The dense capacity control doubles the shared FFN width to 2048,
approximately matching the per-query active FFN capacity of two
experts ($\sim$1.05M FFN parameters per decoder layer, excluding
shared attention blocks). MR2HD uses 8 ROI bins, a 0.1 span expansion
ratio, and scales $\{1,2,4\}$. We train using AdamW~\cite{adamw} with
learning rate and weight decay both $1\times10^{-4}$, batch size 32,
and 250 epochs.
All EviDETR variants are evaluated over three random seeds, and results are
reported as mean $\pm$ standard deviation.

\subsection{Main Results}

\textbf{QVHighlights.}
Table~\ref{tab:qvhighlights} compares EviDETR with representative
temporal grounding methods on QVHighlights using the
CLIP+SlowFast feature setting.

\begin{table*}[t]
\centering
\caption{Comparison on the QVHighlights validation split. All methods use
CLIP+SlowFast features unless otherwise noted. Best and second-best
results are shown in bold and underlined, respectively.}
\label{tab:qvhighlights}

\setlength{\tabcolsep}{4.6pt}
\renewcommand{\arraystretch}{1.05}
\footnotesize

\begin{tabular}{l|cc|ccc|cc}
\hline

\multirow{3}{*}{Model}
& \multicolumn{5}{c|}{Video Moment Retrieval}
& \multicolumn{2}{c}{Highlight Detection} \\

\cline{2-8}

& \multicolumn{2}{c|}{R1}
& \multicolumn{3}{c|}{mAP}
& \multicolumn{2}{c}{$\geq$ Very Good} \\

\cline{2-3}
\cline{4-6}
\cline{7-8}

& @0.5
& @0.7
& @0.5
& @0.75
& Avg.
& mAP
& HIT@1 \\

\hline

UVCOM~\cite{uvcom}
& 65.10 & 51.81
& -- & -- & 45.79
& 40.03 & 63.29 \\

QD-DETR~\cite{qddetr}
& 62.68 & 46.66
& 62.23 & 41.82 & 41.22
& 39.13 & 63.03 \\

TD-DETR~\cite{tddetr}
& 65.88 & \underline{53.67}
& 66.43 & \textbf{49.86} & \textbf{49.05}
& -- & -- \\

LD-DETR~\cite{lddetr}
& 69.01 & 53.19
& \textbf{68.43} & 48.25 & 47.93
& \underline{41.66} & 66.80 \\

CG-DETR~\cite{cgdetr}
& 67.35 & 52.06
& 65.57 & 45.73 & 44.93
& 40.79 & 66.71 \\

BAM-DETR~\cite{bamdetr}
& 65.10 & 51.61
& 65.41 & 48.56 & 47.61
& -- & -- \\

TDP-DETR~\cite{tdpdetr}
& \underline{69.16}
& 53.23
& 68.01
& 48.36
& 47.06
& 41.32
& \underline{67.68} \\

\hline

\textbf{EviDETR (Ours)}
& \textbf{69.29}\std{0.52}
& \textbf{54.77}\std{0.53}
& \underline{68.39}\std{0.31}
& \underline{49.26}\std{0.29}
& \underline{48.41}\std{0.09}
& \textbf{41.83}\std{0.39}
& \textbf{68.33}\std{1.26} \\

\hline
\end{tabular}
\end{table*}

EviDETR achieves 69.29 R1@0.5, 54.77 R1@0.7, and 48.41
Avg.\ mAP for moment retrieval, together with 41.83 HD-mAP and
68.33 HIT@1 for highlight detection. It ranks first on both recall
metrics and both HD metrics, while ranking second on the three mAP
metrics. At the stricter R1@0.7 threshold, it exceeds the second-best
TD-DETR result by 1.10 percentage points, showing consistently strong
performance across retrieval and highlight detection.

\textbf{Evaluation across datasets.}
We further evaluate the model on TACoS and Charades-STA to test
whether the same design transfers across different video domains and
temporal granularities. The results are summarized in
Table~\ref{tab:cross_dataset}.

\begin{table}[t]
\centering
\caption{Cross-dataset results using CLIP+SlowFast features. EviDETR uses
the Charades-STA test split and TACoS test split; baselines follow
their reported splits. Best and second-best results are shown in
bold and underlined, respectively.}
\label{tab:cross_dataset}

\setlength{\tabcolsep}{3.6pt}
\renewcommand{\arraystretch}{1.05}
\footnotesize

\begin{tabular}{l|cc|cc}
\hline

\multirow{2}{*}{Model}
& \multicolumn{2}{c|}{Charades-STA}
& \multicolumn{2}{c}{TACoS} \\

\cline{2-5}

& R1@0.5 & R1@0.7
& R1@0.5 & R1@0.7 \\

\hline

UVCOM~\cite{uvcom}
& 59.25 & 36.64
& 36.39 & 23.32 \\

CG-DETR~\cite{cgdetr}
& 58.40 & 36.30
& 39.50 & 23.40 \\

BAM-DETR~\cite{bamdetr}
& \underline{59.95} & \underline{39.38}
& \underline{41.54} & \underline{26.77} \\

\hline

\textbf{EviDETR (Ours)}
& \textbf{60.28}\std{0.41}
& \textbf{40.13}\std{0.57}
& \textbf{44.09}\std{1.53}
& \textbf{26.94}\std{1.01} \\

\hline
\end{tabular}
\end{table}
On the TACoS test split, EviDETR obtains 44.09 R1@0.5 and
26.94 R1@0.7, exceeding the listed baselines on both metrics.
On the Charades-STA test split, it achieves 60.28 R1@0.5 and
40.13 R1@0.7, also outperforming the listed baselines.

\subsection{Ablation Studies}

We conduct ablation studies on QVHighlights to evaluate the
contribution of the proposed components and the decoder design.
The baseline omits SFR and MR2HD and uses a simple decoder without
multi-stage refinement, auxiliary supervision, or MoE routing.
The results are reported in Table~\ref{tab:ablation}.

\begin{table}[t]
\centering
\caption{Ablation and decoder analysis on the QVHighlights validation split.}
\label{tab:ablation}

\footnotesize
\setlength{\tabcolsep}{2.7pt}
\renewcommand{\arraystretch}{1.04}

\begin{tabular*}{\columnwidth}
{@{\extracolsep{\fill}}lcccc@{}}
\hline
Configuration
& R1@0.5
& R1@0.7
& HD-mAP
& HIT@1 \\
\hline

\multicolumn{5}{l}{\textit{Component ablation}} \\

Baseline
& 52.26\std{0.90}
& 29.61\std{0.85}
& 39.42\std{0.24}
& 62.32\std{0.95} \\

+ SFR
& 55.35\std{0.24}
& 35.61\std{0.32}
& 40.46\std{0.24}
& 63.03\std{0.18} \\

+ SFR + TTop2MoE
& 67.97\std{0.22}
& 52.45\std{0.32}
& 41.12\std{0.38}
& 65.80\std{0.56} \\

+ SFR + TTop2MoE\\ + MR2HD
& \textbf{69.29}\std{0.52}
& \textbf{54.77}\std{0.53}
& \textbf{41.83}\std{0.39}
& \textbf{68.33}\std{1.26} \\

\hline
\multicolumn{5}{l}{\textit{Decoder/active-capacity analysis}} \\

3-stage Shared FFN
& 65.03\std{0.96}
& 46.31\std{0.55}
& 36.20\std{0.35}
& 62.56\std{0.19} \\

3-stage 2$\times$-width FFN
& 68.84\std{0.71}
& 53.68\std{0.43}
& \textbf{42.03}\std{0.10}
& 67.48\std{0.58} \\

3-stage TTop1MoE
& 66.84\std{0.16}
& 51.68\std{0.27}
& 41.05\std{0.17}
& 65.68\std{0.38} \\

\textbf{3-stage TTop2MoE}
& \textbf{69.29}\std{0.52}
& \textbf{54.77}\std{0.53}
& 41.83\std{0.39}
& \textbf{68.33}\std{1.26} \\

\hline
\end{tabular*}

\vspace{2pt}

\raggedright
{\footnotesize
The 2$\times$-width FFN is a shared dense FFN with width 2048 versus
1024 per expert, approximately matching the active FFN capacity of two
routed experts. TTop1MoE and TTop2MoE use the same 8-expert pool. In the
component ablation, TTop2MoE denotes the full three-stage decoder design
with auxiliary supervision and sparse Top-2 routing.
}
\end{table}
\textbf{Ablation and decoder analysis.}
Starting from the baseline, SFR improves R1@0.7 from 29.61 to 35.61,
indicating that explicit query-conditioned reweighting helps temporal
grounding. The full TTop2MoE decoder design then raises R1@0.7 to
52.45; because it also adds three-stage refinement and auxiliary
supervision, this gain reflects the overall decoder design rather than
sparse routing alone. MR2HD further raises R1@0.7 to 54.77 and HIT@1
from 65.80 to 68.33. In the capacity control, doubling the shared FFN
width from 1024 to 2048 raises R1@0.7 from 46.31 to 53.68, showing
that increased capacity explains a substantial part of the decoder gain.
With approximately matched per-query active capacity, TTop2MoE still
reaches 54.77 and improves HIT@1 from 67.48 to 68.33, supporting an
additional benefit beyond simply widening the dense FFN. Top-1 routing
reduces R1@0.7 to 51.68 and HIT@1 to 65.68, further supporting Top-2
multi-expert refinement.

\section{CONCLUSION}
\label{sec:conclusion}
We presented EviDETR, an evidence-preserving framework for joint moment retrieval and highlight detection. EviDETR combines SFR for query-relevant representation enhancement, TTop2MoE for query-adaptive conditional refinement, and MR2HD for explicit retrieval-to-highlight evidence transfer. Experiments on QVHighlights demonstrate strong joint MR/HD performance, while results on TACoS and Charades-STA show consistent effectiveness across different temporal grounding benchmarks. Ablation studies further validate the contributions of the proposed components and decoder design.

\section{Acknowledgment} 
The authors received no specific funding for this work and declare
no relevant financial or nonfinancial interests.

\vfill\pagebreak
\bibliographystyle{IEEEbib}
\bibliography{strings,refs}

\end{document}